\pdfoutput=1

\documentclass[11pt]{article}

\usepackage{acl}

\usepackage[title]{appendix}
\usepackage{times}
\usepackage{latexsym}
\usepackage{graphicx} 

\usepackage[T1]{fontenc}

\usepackage[utf8]{inputenc}

\usepackage{microtype}

\usepackage{times}
\usepackage{latexsym}
\usepackage{subcaption}
\usepackage{multirow,bigdelim}
\usepackage{amsmath}
\usepackage{rotating}
\usepackage{booktabs,xcolor}
\usepackage{xspace}
\usepackage{booktabs}
\usepackage{multirow}
\usepackage{todonotes}
\usepackage{url}
\usepackage{enumitem}

\newcommand{\cons}{\texttt{CONS}\xspace}
\newcommand{\incons}{\texttt{INCONS}\xspace}
\newcommand{\constest}{\texttt{CONSTest}\xspace}
\newcommand{\chextrain}{\texttt{CheXTrain}\xspace}
\newcommand{\chexval}{\texttt{CheXVal}\xspace}

\newcommand{\negincons}{\texttt{NegINCONS}\xspace}
\newcommand{\chexinconstest}{\texttt{CheXINCONSTest}\xspace}
\newcommand{\neginconstest}{\texttt{NegINCONSTest}\xspace}

\title{Modeling Disagreement in Automatic Data Labelling for Semi-Supervised Learning in Clinical Natural Language Processing}

\author{Hongshu Liu$^1$\thanks{~Equal contribution.}\quad Nabeel Seedat$^{2^*}$ \quad Julia Ive$^3$\\ 
$^1$ Imperial College London \\
$^2$ University of Cambridge \\
$^3$ Queen Mary University of London \\
hongshu.liu19@imperial.ac.uk, ~ ns741@cam.ac.uk, ~j.ive@qmul.ac.uk}

\begin{document}
\maketitle
\begin{abstract}
  Computational models providing accurate estimates of their uncertainty are crucial for risk management associated with decision making in healthcare contexts. This is especially true since many state-of-the-art systems are trained using the data which has been labelled automatically (self-supervised mode) and tend to overfit. In this work, we investigate the quality of uncertainty estimates from a range of current state-of-the-art predictive models applied to the problem of observation detection in radiology reports. This problem remains understudied for Natural Language Processing in the healthcare domain. We demonstrate that Gaussian Processes (GPs) provide superior performance in quantifying the risks of 3 uncertainty labels based on the negative log predictive probability (NLPP) evaluation metric and mean maximum predicted confidence levels (MMPCL), whilst retaining strong predictive performance. 
\end{abstract}

\section{Introduction} 
While current machine learning and deep learning models have shown great success in a variety of automated classification tasks~\cite{Rajkomar2018,Yang2018}, sensitive healthcare contexts, necessitate that these models quantify the risks of different diagnosis decisions, i.e. \emph{positive} - the illness is highly likely, \emph{uncertain} - the illness is likely, with lower risks, \emph{negative} - the illness is highly unlikely.  The 3 uncertainty labels (also our class labels) allow each diagnosis decision to be made, by factoring in the degrees of uncertainty.  

This is especially relevant as human annotation is very costly and model training data is often labelled by automatic labellers that use ontologies or are rule-based. For example, the UMLS (Unified Medical Language System) ontology \cite{bodenreider2004unified} is almost predominantly used to match linguistics patterns in clinical text to medical concepts (e.g. using the MetaMap tool ~\cite{aronson2006metamap}). The corpora annotated in this fashion are used to learn neural detectors of medical concepts in the self-supervised setting~\cite{zhang2021selfsupervised}. Rule-based automatic labellers such as CheXpert \cite{irvin2019chexpert}, which builds on NegBio \cite{peng2018negbio} use rules predefined by experts to extract clinical observations from the free text of radiology reports. Both use different rules, where CheXpert is more conservative, with more uncertain labels.

Most clinical Natural Language Processing (NLP) models focus on optimizing predictive classification performance on class labels with point estimates, but pay less attention to further quantify the risks and uncertainty of these labels.  Bayesian methods offer as a solution, a principled approach to produce well-calibrated uncertainty estimates. Popular Bayesian approximations (e.g., model ensembles, dropout techniques etc) have primarily been studied in the general NLP domain~\cite{trustmodeluncertainty} or for continuous healthcare data ~\cite{uncertaintymetrics,Leibig:2017aa}. More rarely, those studies are in Clinical Natural Language Processing (NLP)~\cite{Yang_2019_CVPR_Workshops,Guo2021}, which has a unique asymmetric risk where underestimates and overestimates of confidence should be evaluated differently.

In this work, we build a model using Gaussian Processes (GPs)~{\cite{williams2006gaussian}}, a Bayesian non-parametric method to quantify the uncertainty and disagreements associated with automatic data labels.  Instead of learning distributions over model parameters as in Bayesian Deep Learning (BDL), GP models are non-parametric and learn distributions over functions, thus allowing for better generalisation.

We compare GP performance to two types of Bayesian approximations: frequentist ensembles using both Random Forests (RFs)~\cite{Breiman:2001:RF} and Deep Ensembles (ENS)~\cite{ens}, which are considered state-of-the-art for uncertainty estimation in predictive settings \cite{lobacheva2020power,seedat2019towards}. We make use of the three automatic uncertainty labels (considered noisy labels), as provided by the CheXpert and Negbio labellers for diagnostics using radiology reports and make the following contributions:

\noindent 
(1) We study the quality of uncertainty estimates of a new range of uncertainty-aware models, previously understudied in clinical NLP;\\
(2) We successfully apply Sparse Gaussian Processes (GPs) to the diagnosis detection task in radiology reports. To the best of our knowledge, this is the first application in clinical NLP and we demonstrate that GPs provides superior uncertainty representations yet retain comparable predictive performance when compared to baselines Random Forests and Deep Ensembles;\\
(3) We demonstrate the utility of uncertainty estimates for the models trained with data labelled automatically (self-supervised) with disagreements (differences of opinions from rule-based CheXpert and Negbio labellers). GPs outperform the RF and ENS models: providing less certainty in the predictions for the test cases where labels assigned by different labellers disagree. Correct identification of such cases is crucial for reliable and trustworthy risk management, where any uncertain cases are referred for screening or to a human expert.

\section{Methodology}

Gaussian Processes (GPs) offer a principled probabilistic modelling approach and provide uncertainty estimates without needing post-processing, hence motivating our modelling decision. The GP is defined by GP($\mu(\textbf{x})$, $\textbf{K}(\textbf{x},\textbf{x}^{'},\theta)$), a mean function and covariance function respectively. When the GP is realized on observed data it is given as: $p(\textbf{y}|\textbf{f}) =  \prod_{i=1}^{n} p(\textbf{y}_{i})|\textbf{f}_{i})$. where $\textbf{f}$ is a vector of latent functions $ p(\textbf{f}|\textbf{x},\theta)= \prod_{i=1}^{n} \mathcal{N}(\textbf{f}_{i}; \textbf{0}, \textbf{K}_{j})$. Due to space limitations we refer to~\citet{williams2006gaussian} for more details. In our work, we apply GPs to the classification task of clinical observation detection. The challenge with using GPs for classification is that an approximation for the posterior is required.\footnote{Since the likelihood can no longer be assumed Gaussian due to the discrete labels, computing the posterior becomes computationally intractable and hence, requires methods such as Markov Chain Monte Carlo and Variational Inference to approximate the posterior.} GPs also have a complexity of $O(n^3)$ \cite{williams2006gaussian}, due to the matrix inversion, making it computationally intensive for NLP tasks using large datasets and high dimensional embeddings. 

Hence, in dealing with the these challenges, we use Sparse GPs with Black-box variational inference \cite{black-box2015} (minimize the log-evidence lower bound) and use the Reparameterization Trick \cite{kingma2014autoencoding} to approximate the GPs - in a more computationally feasible method, using a small set of latent inducing points.  This approach is based on~\citet{titsias2009variational}.  We also use automatic relevance determination (ARD) \cite{MacKay1996} to allow for more flexibility in our kernel representation. 

\section{Experimental settings}

\subsection{Data}
 
We study free-text radiology reports in the MIMIC-CXR database v2.0.0 \cite{johnson2019mimiccxrjpg}. The dataset is labelled by both the CheXpert (primary label) and Negbio labellers. CheXpert is based on Negbio, but they follow different strategies in mention detection: Negbio uses MetaMap and CheXpert curates concepts predefined by clinical experts. 

We analyse the `Edema' pathology- 65,833 reports with 3 uncertainty labels: \emph{positive}, \emph{uncertain} and \emph{negative} as explained earlier. Edema was chosen due to the large data size, more inconsistent labels between the two labellers and more balanced split between classes. We partition the `Edema' examples into: (1) 63,482 \underline{consistent} labels ($\sim$96\% dataset), where both CheXpert and Negbio labels agree (\cons : 26,455 - \emph{positive}, 11,781 - \emph{uncertain}, 25,246 - \emph{negative}) and (2) 2,351 \underline{inconsistent} labels making up $\sim$4\% of the total dataset, where CheXpert and Negbio labels disagree (\incons, 522: \emph{positive}, 1,317 - \emph{uncertain}, 512 - \emph{negative}). Validation and Test sets preserve the described class proportions and contain 10\% of the dataset each.

In our setup, we use two variants of the inconsistent test sets (i.e. labellers disagree) to quantify the generalisation to different labeling mechanisms namely: \neginconstest~and \chexinconstest, taking as the `ground truth', the automatic Negbio and CheXpert labels, respectively. 
CheXpert is considered the primary label, hence the Train and Validation sets only use CheXpert labels, also containing both consistent and inconsistent data as mentioned above.

Text pre-processing (tokenisation, lower-casing, white space and punctuation removal) is done using \texttt{Texthero}. The average token length is $\sim$ 43.5 tokens. Each training example's input data is represented by a 200 dimensional fixed length vector averaged over tokens. We use the biomedical \texttt{Word2Vec-Pubmed} word embeddings ~\cite{chiu2016train} to represent tokens. 

\begin{table}[h]
    \centering
        \caption{Results comparing GPs, RF and ENS based on accuracy and NLPP for both inconsistent (OOD) and consistent test sets.  Bold highlights the best results.}
\scalebox{0.6}{
\begin{tabular}{c@{\hspace{1cm}}cccccccc}
    \toprule
& \multicolumn{2}{c}{\neginconstest} & \multicolumn{2}{c}{\chexinconstest} & \multicolumn{2}{c}{\constest} & \\ 
\cline{2-7}
         & Acc$\uparrow$ & NLPP$\downarrow$ & Acc$\uparrow$ & NLPP$\downarrow$ &  Acc$\uparrow $& NLPP$\downarrow$ \\
\midrule
   \bf GP & \bf 0.294 & \bf 1.234 & 0.519 & 0.634 &   0.816 &  0.293 \\
\midrule
    RF & 0.271 & 1.451  & \bf 0.579 & \bf 0.598 & 0.795 & 0.316 \\
\midrule
    ENS  & 0.263 & 1.281  & 0.468 &  0.691 & \bf 0.840 & \bf 0.269 \\
  \bottomrule
\end{tabular}}
    \label{perf}
\vspace{-5mm}
\end{table}

\subsection{Models}

Our Sparse GP model consists of 300 inducing points and a radial basis function (RBF) kernel making use of ARD to learn the length scales. 
The expected likelihood term in $L_{elbo}$ is estimated using Monte Carlo sampling. The GP model is trained stochastically with RMSProp (learning rate = 0.003, epochs=2, batch size=500).

The RF model (using \texttt{Scikit-learn} \cite{scikit-learn}) has the following hyper-parameters based on random search over standard parameters: 300 trees and a max depth of 40. Isotonic regression is used to obtain well-calibrated predictive distributions (comparable to the GP).

The Deep Ensemble (ENS) model based on ~\cite{ens} has: 5 randomly initialized models (MLP with 3-hidden layers with 200 hidden units/layer (with batch norm), Adam (learning rate = 3e-3), epochs=10)). Following~\cite{ens}, we use adversarial training (fast gradient sign method \cite{adv}), for better model robustness. 

\subsection{Evaluation metrics}
 The models are evaluated based on predictive performance (accuracy) and we quantify the quality of the uncertainty representations using the negative log predictive probability $NLPP= -log~p(C=y_{n}|x_{n})$. NLPP penalises both overconfident predictions that are incorrect predictions and under-confident predictions that are correct. We optimise for higher predictive performance and lower NLPP.  
 
Since we do not have ground truth labels for multiple human experts, we additionally assess the mean maximum predicted confidence level (MMPCL) of a certain test set (see Table 2),  $ MMPCL =  \frac{1}{N}\sum_{n=1}^{N} max_{j}~p(y_{n}=C_{j}|x_{n})$, where N is total number of data points in the specific test set and $C_{j}$ is per class label. 

\section{Results}

We carry out two experiments:
(1) We measure the performance in uncertainty representation and predictive performance of the model trained and validated on primary labels (CheXpert - \chextrain and \chexval, both contain consistent and inconsistent dataset) 
for the inconsistent Test data (disagreement- \neginconstest, \chexinconstest), as well as for the consistent Test data (agreement - \constest); 
(2) We conduct a group-wise performance analysis, particularly assessing how different models deal with the asymmetric risk of false negatives. 

\subsection{Performance across labeller agreement}

\textbf{Inconsistent Labels.} Evaluating performance on inconsistent labels (CheXpert and Negbio disagree) highlights the use case of evaluation on out-of-distribution (OOD) labels (since the models are trained on CheXpert labels, but tested on Negbio labels - \neginconstest). Table \ref{perf} illustrates the average predictive accuracy and average NLPP for all the test sets (\neginconstest, \chexinconstest, \constest). Overall, all models exhibit a higher NLPP across on both inconsistent test sets compared to \constest. The high NLPP highlights the models are able to correctly represent predictive uncertainty, when there was in fact disagreement in predictions.

For experiments, comparing \neginconstest and \chexinconstest, we note that CheXpert has more conservative labels compared to Negbio: i.e  \chexinconstest~ contains 63\% of uncertain labels and 13\% negative, whilst \neginconstest~ 30\% uncertain and 63\% negative. When the GP is evaluated on \neginconstest (labels are OOD) it outperforms the RF, yet the RF outperforms the GP on \chexinconstest (labels are in-distribution to training labels - CheXpert). In the case of ENS, it outperforms GP when evaluated on \constest but performs worse on inconsistent test sets (both \chexinconstest and \neginconstest). These results highlight the RF can be biased/overfit the ``conservative'' distribution of the training data, whereas the GP provides more `neutral' predictions, generalising better to OOD data (e.g. \neginconstest).  The poor performance of ENS  on inconsistent test data (\chexinconstest (in-distribution) ~, \neginconstest (OOD)), highlights that ENS does not generalize to OOD data nor does it handle noisy labels well.  
The GPs generalization performance is highly desirable in medical settings and our results on GP generalization is corroborated by \cite{schulam2017reliable}. For further analysis based on calibration see Appendix \ref{calibration}.

\textbf{Consistent Labels.}
As a control experiment and to account for the imbalanced data, we randomly sample the \cons~test set (\constest) such that it has the same size as the other two inconsistent test sets. All models show lower NLPP compared to the inconsistent experiments as the cases within \constest tend to have high certainty/agreement.

\subsection{Group-wise analysis for asymmetric risk}

A significant asymmetric risk in the healthcare context lies in False Negatives (FN), i.e. missing a positive diagnosis. Ideally, models should represent FNs with low predictive confidence, whilst True Positives (TPs) should have high predictive confidence. As shown in Table~\ref{conf}, we compare the MMPCLs of FNs and TPs for the \emph{positive} and \emph{uncertain} labels for \neginconstest~ and \constest and ignore the \emph{negative} labels.

On \constest (labellers agree), the GP provides lower MMPCL for FN predictions and higher MMPCL for TP predictions when compared to the RF across all test sets. Whilst, ENS has good performance for \underline{FN \emph{positive}} and \underline{TP \emph{uncertain}}, this is traded off for reduced performance for \underline{FN \emph{uncertain}} and \underline{TP \emph{positive}}.  We note that since GP has better balanced performance across both classes (positive and uncertain) it is likely to generalise better/not overfit, thus dealing with asymmetric risk in a more appropriate and balanced manner.

Similar patterns are observed for \negincons, with the GP better representing the asymmetric risk (for both positive and uncertain) when compared to both ENS and RF. However, the results are less conclusive due to the sparsity of FNs: we have only one FN data point with \emph{positive} label for GP and RF. Further, the two data points are different, hence the result where the RF outperforms on this subset, cannot be considered as reliably representative. That said, the absence of \underline{TP \emph{uncertain}} labels for RF model mirrors the disagreements between the two labellers and highlights the inherent difficulty in predicting uncertain cases.

Overall, the GP and ENS give similar performance for in-distribution test data \constest, however on the OOD test data \neginconstest, the GP gives better performance.

\begin{table}[h]
    \centering
        \caption{Results for GP, RF \& ENS, wrt. MMPCL for FNs and TPs  evaluated on \neginconstest~ and \constest. Bold highlights the best results: lower confidence FN and higher confidence TP is desirable.}
\scalebox{0.7}{
\begin{tabular}{llc@{\hspace{1cm}}ll@{\hspace{0.5cm}}ll}
    \toprule
&  & \multicolumn{2}{c}{$\neginconstest$} 
& \multicolumn{2}{c}{$\constest$} \\ 
\cline{2-6}
  & & Positive  & Uncertain & Positive & Uncertain   \\
\midrule
\multirow{2}{*}{\rotatebox{90}{\textbf{FN}}}
  &  GP   & 0.565 & \textbf{0.800}  & 0.647 & \textbf{0.638}\\
    
  &  RF     & \textbf{0.408} & 0.857  & 0.657 & 0.661 \\
 &  ENS     & 0.694 & 0.876   & \textbf{0.624} & 0.640 \\
    
\midrule
\midrule
\multirow{2}{*}{\rotatebox{90}{\textbf{TP}}}
&    GP   & \textbf{0.711} & \textbf{0.605}  & \textbf{0.797} & 0.752 \\
    
&    RF  & 0.680 & 0  & 0.774 & 0.730 \\
  &  ENS    & 0.670 & 0.318  & 0.780 & \textbf{0.834} \\
  \bottomrule
\end{tabular}}
    \label{conf}
 \vspace{-5mm}
\end{table}

\section{Conclusions}

We present a Sparse GP model for clinical NLP that is capable of quantifying the risks of different uncertainty labels. This model can be particularly advantageous in self-supervised learning scenarios with automatically labelled datasets. It predicts lower confidence estimates for the test examples where different labelling heuristics tend to disagree. Also, overall our sparse GP provides more conservative predictions by showing lower confidence in incorrect predictions. This is important in healthcare contexts where potential harms from a model error can be very extensive as compared to singular errors of a doctor. This also implies that our model could be used to flag clinical notes for further review (e.g. by a human expert).

Whilst, this work is focused on uncertainty in labelling for medical tasks, we believe it highlights the utility of probabilistic models applied to ``noisy'' labels and that similar methods could provide utility for NLP based automated labelling tasks. Furthermore, we hope that this work will spur further research in this understudied area of automated labelling in clinical NLP - where annotated data is limited and expensive.

\section{Ethical Considerations}

This study has been conducted strictly observing relevant guidelines for usage of the MIMIC III data. Medical decision making has high-stakes consequences, hence to mitigate any potential risks we note that the presented models should be used only to assist human medical experts in their decisions. Before the deployment of these models, they should be to subject to systematic validation.

\bibliography{main}

\newpage
\onecolumn

\begin{appendices}

\section{Calibration.}\label{calibration} 
\begin{figure}[h]
  \centering
  \includegraphics[width=0.75\hsize]{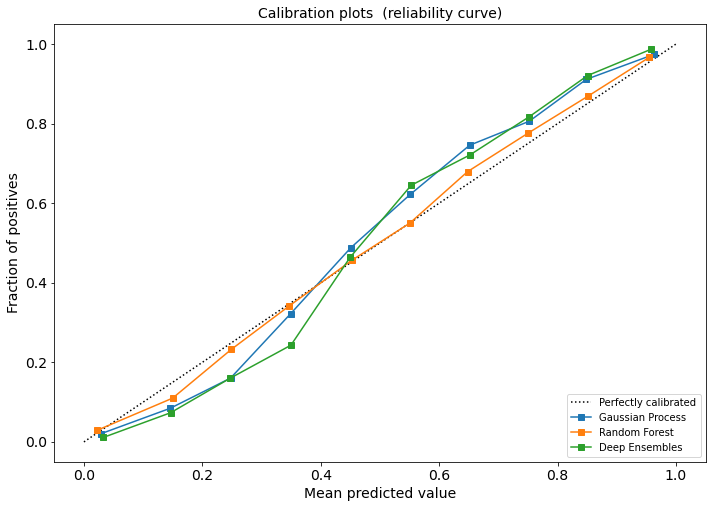}
  \caption{Reliability diagram for the \constest
set: fraction of positives (y-axis) vs mean predicted value (x-axis)}
  \label{fig:reliability}
  \vspace{-5mm}
\end{figure}

Figure \ref{fig:reliability} compares the calibration of the GP, RF and ENS models, which all generate probabilities close to the optimal curve. While, seemingly the calibrated RF model (using Isotonic Regression) gives more optimal predictions than the GP and ENS models, this result is based on the \cons ~test set (which is the same as the training distribution where both CheXpert and NegBio labels agree). However, both the GP and ENS are slightly less optimal as they generalize better across labellers and hence the GP and ENS overfit less to \cons ~ specifically, when compared to RF.

\section{Examples of False Negatives}
\label{appsec:examples}
A data point from consistent test set labelled \textit{Positive} classified as \textit{Negative}, (paraphrased)

\vspace{3ex}

\noindent\fbox{%
    \parbox{\textwidth}{%
    AGAIN MILD PROMINENCE OF THE PULMONARY INTERSTITIAL MARKINGS \textbf{SUGGESTIVE OF PULMONARY EDEMA}, STABLE. SUBSEGMENTAL ATELECTASIS AT THE LUNG BASES. THERE IS NO DEFINITE CONSOLIDATION. NO PNEUMOTHORACES ARE SEEN.
    }%
}

\vspace{3ex}

\noindent A data point from inconsistent test set labelled\\ \textit{Positive} classified as \textit{Negative}, (paraphrased)

\vspace{3ex}

\noindent\fbox{%
    \parbox{\textwidth}{%
    THE PATIENT IS STATUS POST DUAL-LEAD LEFT-SIDED AICD WITH LEADS EXTENDING TO THE EXPECTED POSITIONS OF THE RIGHT ATRIUM AND RIGHT VENTRICLE. PATIENT IS STATUS POST MEDIAN STERNOTOMY. THERE IS PROMINENCE OF THE CENTRAL PULMONARY VASCULATURE \textbf{SUGGESTING MILD EDEMA}/VASCULAR CONGESTION. THE CARDIAC SILHOUETTE REMAINS QUITE ENLARGED. DIFFICULT TO EXCLUDE SMALL PLEURAL EFFUSIONS. THERE IS BIBASILAR ATELECTASIS. NO DISCRETE FOCAL CONSOLIDATION IS SEEN, ALTHOUGH OPACITY AT THE LUNG BASES PROJECTED ON THE LATERAL VIEW WHILE MAY BE DUE TO OVERLYING SOFT TISSUES, CONSOLIDATION IS DIFFICULT TO EXCLUDE.
    }%
}
\end{appendices}

\end{document}